# Classifiers of Data Sharing Statements in Clinical Trial Records

Saber JELODARI MAMAGHANI [a], Cosima STRANTZ [a] and
Dennis TODDENROTH [a,1]
[a] *Medical Informatics, University Erlangen-Nuremberg, Germany*

**Abstract.** Digital individual participant data (IPD) from clinical trials are increasingly distributed for potential scientific reuse. The identification of available IPD, however, requires interpretations of textual data-sharing statements (DSS) in large databases. Recent advancements in computational linguistics include pre-trained language models that promise to simplify the implementation of effective classifiers based on textual inputs. In a subset of 5,000 textual DSS from ClinicalTrials.gov, we evaluate how well classifiers based on domain-specific pre-trained language models reproduce original availability categories as well as manually annotated labels. Typical metrics indicate that classifiers that predicted manual annotations outperformed those that learned to output the original availability categories. This suggests that the textual DSS descriptions contain applicable information that the availability categories do not, and that such classifiers could thus aid the automatic identification of available IPD in large trial databases.

## 1. Introduction

While clinical research has for some time aggregated findings from previous clinical trials, today the increasing availability of digital individual participant data (IPD) yields increasing opportunities to synthesize new medical knowledge. Effective data reuse, however, requires that collected IPD should be discoverable and accessible to other researchers for further analysis [1]. Online platforms designed to share patient-level trial data promote transparency [2] but may introduce additional requirements that can impede data reuse, especially when access criteria remain *'scattered across multiple web pages'* or *'buried within legal agreements'* [3]. Previous investigations have indicated that only a minority of clinical trials and observational studies declare clear intentions to share IPD, and even then vital details for data access may be unclear [4]. These previous investigations relied on the manual analysis of data-sharing statements (DSS) in trial registries like ClinicalTrials.gov, which has requested categorical and textual information about data-sharing intentions from trialists since 2018.

As of December 2023, ClinicalTrials.gov contains records for more than 470,000 clinical trials, highlighting the extensive collection of medical research data that is potentially accessible for study. This large quantity of records impedes exhaustive manual analyses of DSS. Natural Language Processing (NLP) methods that have been developed in the last years, such as pre-trained models, can potentially process and categorize such extensive textual datasets. Modern language models employ a transformer architecture with a so-called self-attention mechanism to model the relationship between sequential elements of input texts [5]. Transformers use encoders and decoders to comprehend input data for tasks such as text classification or the production of textual outputs.

---

[1] Corresponding Author: Dennis TODDENROTH; E-mail: dennis.toddenroth@fau.de.

One example of an encoder-based language model is Bidirectional Encoder Representations from Transformers (BERT), which models textual contexts from previous as well as subsequent tokens, and is suitable for text categorization tasks [6]. Here we aim to evaluate the effectiveness of BERT-based encoder models for the categorization of DSS extracted from clinical trial descriptions.

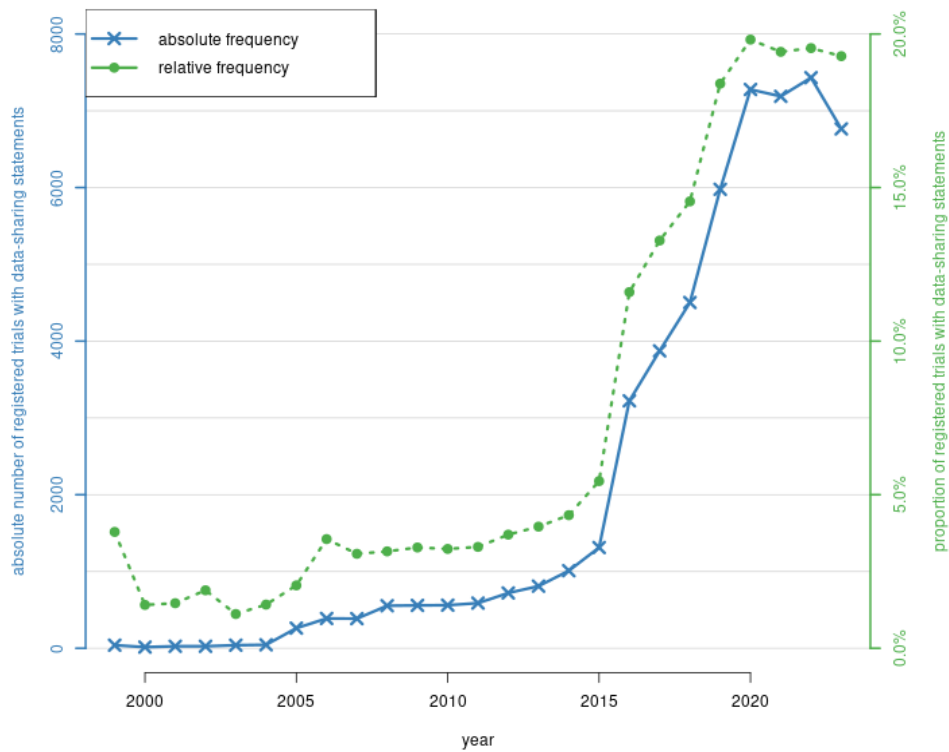

**Figure 1.** Number of yearly clinical trials for which data-sharing statements are registered at ClinicalTrials.gov, queried via its application programming interface.

## 2. Methods

In order to evaluate the effectiveness of modern language models for DSS classification, we focused on three BERT versions that have been pre-trained using domain-specific corpora. SciBERT has been trained on a compilation of scientific papers [7]. BioBERT has been designed to process biomedical texts [8]. BlueBERT has been trained on a combination of biomedical and clinical materials [9]. The selection of these models was based on their proposed efficacy for processing specialist language.

Table 1. Examples of availability categories from ClinicalTrials.gov, DSS, and manually annotated labels.

| Trial identification | Original availability category | DSS | Manually annotated label |
|---|---|---|---|
| NCT03822728 | No | The investigators will make our participant data available to other researchers after completion of this study | Yes |
| NCT03463993 | Yes | It is undecided whether the IPD will be available to other researchers. Clearance is required first from ethical bodies and supervisors | Undecided |
| NCT03288623 | Undecided | De-identified individual participant data for all primary and secondary outcome measures will be made available | Yes |

For this analysis, we queried a dataset from ClinicalTrials.gov via its application programming interface (API), and converted categorical IPD sharing status and textual descriptions into comma-separated values (CSV). The extracted dataset initially consisted of 472,608 records, but after data cleaning and excluding records without IPD sharing statements, it was reduced to 30,495 records. This preprocessing included eliminating entries that had duplicates, or that consisted of fewer than 10 characters, and standardizing the data by deleting some uncommon characters and unwieldy phrases (@@,*,'gsk and wrair', 'glaxosmithkline', 'n/a - phase i study').

The original categorical attribute on IPD availability assumed either 'Yes', 'No', or 'Undecided'. Evaluating exemplary trial records demonstrated frequent discrepancies between provided categorical labels and textual DSS descriptions. Assuming that the textual descriptions contain valuable information that the original categories do not reflect, we manually annotated a subset of 5,000 trial-specific DSS, and defined alternative IPD categories, resulting in 2,441 labels classified as "Yes", 1,232 as "No", and 1,327 as "Undecided". Table 1 contains examples of the IPD sharing descriptions with discrepant IPD categories as well as our complementary labels.

We then evaluated how well classifiers based on the mentioned BERT-type language models could learn to reproduce either the original IPD categories or our manual labels from the textual DSS. This aimed to assess the models' capacity to appropriately categorize IPD sharing descriptions.

Following dataset refinement, we split the dataset into training, validation, and testing segments with respective proportions of 70%, 15%, and 15%. The division was implemented to mitigate overfitting through an early stopping mechanism that monitored performance on the validation set, thereby ensuring our models appropriately represented unseen data. Optimization was conducted over 6 epochs using the *AdamW* optimizer, ensuring the models were optimized for our specific dataset. This technical setup as well as the annotated DSS labels are available at *https://github.com/sjelodari/ClinicalTrialIPDClassifier/*.

## 3. Results

Table 2 demonstrates the performance metrics for SciBERT, BioBERT, and BlueBERT, with comparable accuracy rates of approximately 69% for the original IPD categories and 83% for our complemented annotations across the models. Among the evaluated models, SciBERT moderately outperformed the alternatives. An analysis of label agreement with the original IPD categories revealed that 3,130 of 5,000 (62.2%) of our annotated labels agreed with the original three-level IPD categorization. The referenced repository provides additional statistical details such as model-specific confusion matrices.

**Table 2.** Performance metrics of DSS classifiers based on BioBERT, BlueBERT, and SciBERT

| Dataset-Labels | Original availability categories | | | Manually annotated labels | | |
|---|---|---|---|---|---|---|
| | Accuracy | Precision | F1 Score | Accuracy | Precision | F1 Score |
| **SciBERT** | 0.693 | 0.694 | 0.691 | 0.833 | 0.836 | 0.835 |
| **BioBERT** | 0.693 | 0.692 | 0.691 | 0.831 | 0.833 | 0.832 |
| **BlueBERT** | 0.679 | 0.675 | 0.676 | 0.831 | 0.835 | 0.832 |

## 4. Discussion

Our results indicate that the NLP-based classifier reproduced annotated labels better than the original categorical entries. This suggests that there is valuable information in the textual DSS entries that the provided IPD categories do not reflect. Currently, ClinicalTrials.gov does not have a mechanism for filtering trials based on IPD sharing status. We assume that discrepancies between textual DSS and the original availability categories could undermine the potential utility of such a filtering function at the moment. Our findings suggest that pre-trained language models may be effective for the extraction of relevant information from textual DSS, and if implemented in search functions or filtering options could aid the discovery of available IPD.

With the growing volume of registered trials at ClinicalTrials.gov, the capability of specialized models such as SciBERT to assign interpretable categorical labels to descriptions becomes potentially valuable. Its observed effectiveness highlights a potential pathway to improving the discoverability and reusability of IPD records for an increasingly automated form of medical research based on digital artifacts.

As a limitation, we have to concede that our analysis was restricted to DSS that could be exported from clinicaltrials.gov. In practice, many statements include links to additional web-based platforms or direct users to contact the respective research teams to ask for IPD access. This dependency on other resources curtails our ability to assess whether IPD is truly shared, specifically since real data sharing may fall short of declarations [10]. We propose that automated DSS interpretations, however, could support subsequent studies of the relationship between stated intentions and actual IPD sharing. Future methodological research might investigate whether larger pre-trained language models that support prompting instead of learning from examples can lead to an improved classification performance.

## 5. Conclusions

Our findings highlight that the evaluated language models offer consistent performance in classifying IPD sharing statements, with SciBERT slightly outperforming BioBERT and BlueBERT. Trained classifiers might facilitate data discovery and improve accessibility on platforms such as ClinicalTrials.gov, which currently lacks specific search features for IPD status. Improved accuracy in automatic labeling could support researchers in efficiently identifying datasets that align with their study goals.